\documentclass[conference]{IEEEtran}
\IEEEoverridecommandlockouts
\usepackage{cite}
\usepackage{amsmath,amssymb,amsfonts}
\usepackage{algorithmic}
\usepackage{graphicx}
\usepackage{textcomp}
\usepackage{xcolor}
\usepackage{float}
\usepackage{tabularx}
\usepackage{balance}
\usepackage{microtype}
\usepackage[italic]{mathastext}

\usepackage[final]{changes}
\newcommand{\rp}[2]{\replaced{#1}{#2}}

\usepackage{todonotes}

\usepackage{tikz}
\usepackage{verbatim}
\usetikzlibrary{trees}

\def\BibTeX{{\rm B\kern-.05em{\sc i\kern-.025em b}\kern-.08em
    T\kern-.1667em\lower.7ex\hbox{E}\kern-.125emX}}

\newcommand{\featgroup}[1]{\vspace{.0526in}\noindent\emph{#1} \textemdash }

\let\url\nolinkurl

\begin{document}

\title{Characterizing Input Methods \\for Human-to-robot Demonstrations
\thanks{This research was supported in part by NSF award 1830242 and a UW2020 award from the University of Wisconsin--Madison Office of the  Vice Chancellor for Research and Graduate Education. This is a pre-print of the paper published in the Proceedings of the \textit{2019 ACM/IEEE International Conference on Human-Robot Interaction (HRI)}.}
}

\author{\IEEEauthorblockN{Pragathi Praveena,$^1$ Guru Subramani,$^2$ Bilge Mutlu,$^1$ Michael Gleicher$^1$}
\IEEEauthorblockA{$^1$Department of Computer Sciences \ $^2$Department of Mechanical Engineering \\
University of Wisconsin--Madison, Madison, Wisconsin, USA \\
$^1$\{pragathi,bilge,gleicher\}@cs.wisc.edu, $^2$gsubramani@wisc.edu}

}

\maketitle

\begin{abstract}

\rp{Human}{\emph{Human-to-robot}} demonstrations are important in a range of \rp{robotics applications}{\emph{use cases}}, and \rp{are}{can be} created with a variety of input methods.
However, the \rp{design space}{space of designs} for these input methods has not been \rp{extensively studied}{well characterized}.
In this paper, \rp{focusing on demonstrations of hand-scale object manipulation tasks to robot arms with two-finger grippers, we identify distinct \textit{usage paradigms} in robotics that utilize human-to-robot demonstrations, extract abstract \textit{features} that form a design space for input methods, and characterize existing input methods as well as a novel input method that we introduce, the \textit{instrumented tongs}.}{we analyze a range of use cases for human task demonstrations to identify desirable features for input methods. These features make up a design space that is used to assess existing input methods and to develop design specifications for novel methods. We introduce a new input method, the \textit{instrumented tongs}, that meets these specifications.} \rp{We detail the design specifications for our method and}{We} present a user study that compares \rp{it}{our method} against three common input methods: free-hand manipulation, kinesthetic \rp{guidance}{demonstrations}, and teleoperation. Study results show that instrumented tongs provide high quality demonstrations and a positive\rp{}{ user} experience for the demonstrator while offering good correspondence to the target robot. 
\end{abstract}


\newcommand{\figframework}{
\begin{figure}[!t]
  \centering
  \includegraphics[width=\columnwidth]{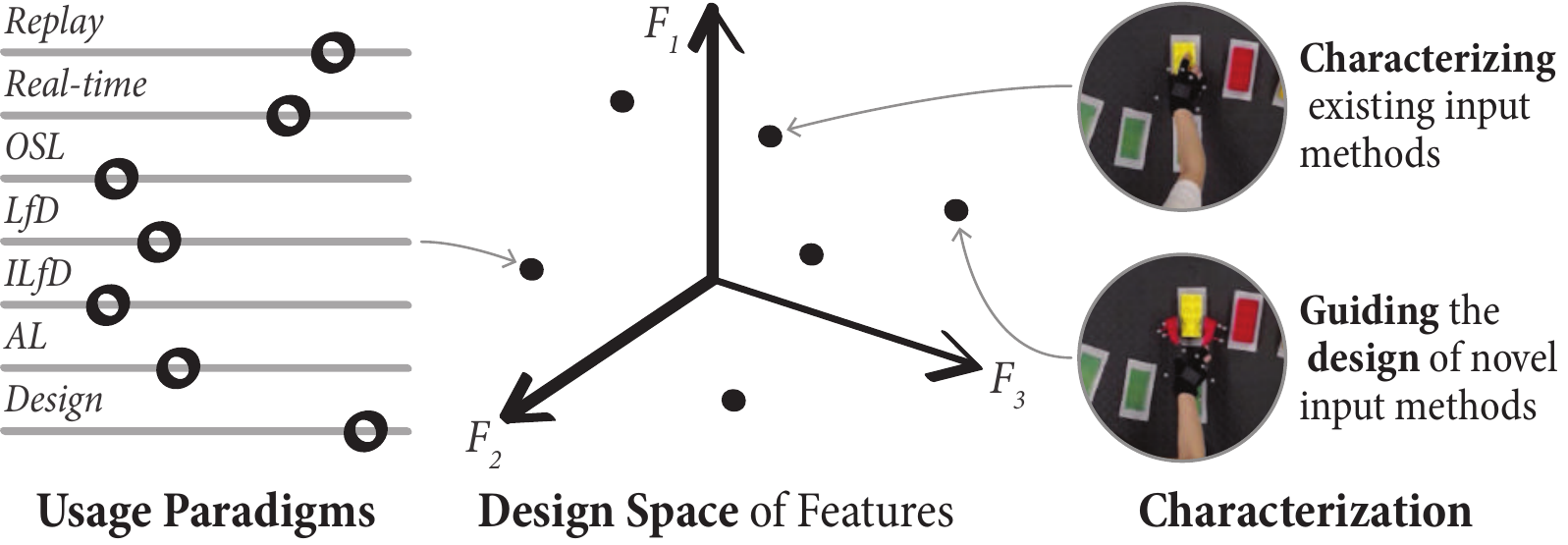}
  \caption{Our framework and its three components: usage paradigms, design space of features, and characterization of input methods.}
  \label{fig:framework}
  \vspace{-3mm}
\end{figure}
}

\newcommand{\figdesign}{
\begin{figure}[!t]
  \centering 
  \includegraphics[width=0.95\linewidth]{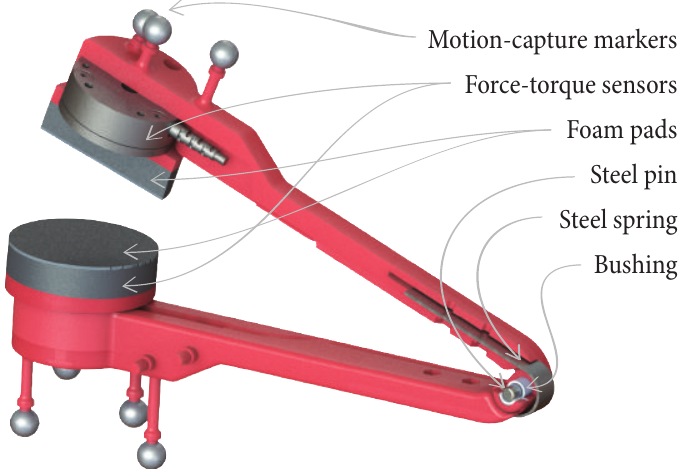}
  \caption{Design of the current prototype of \textit{instrumented tongs}.
  \label{fig:tongsdesign}}
  \vspace{-3mm}
\end{figure} 
}

\newcommand{\figteaser}{
\begin{figure}[!t]
  \centering
  \includegraphics[width=1\linewidth]{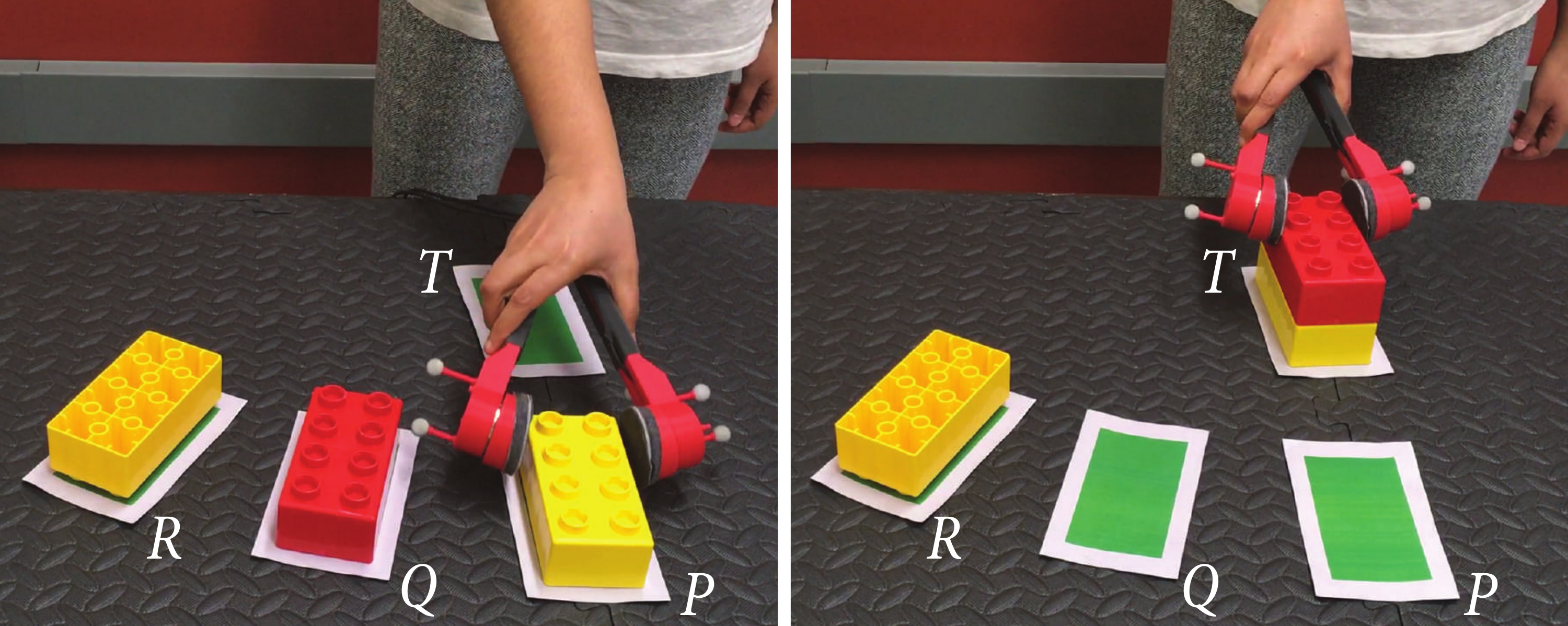}
  \caption{Using \textit{instrumented tongs} to demonstrate Lego block stacking.}
  \label{fig:teaser}
  \vspace{-3mm}
\end{figure}
}

\newcommand{\figall}{
\begin{figure}[!t]
  \centering
  \includegraphics[width=1\linewidth]{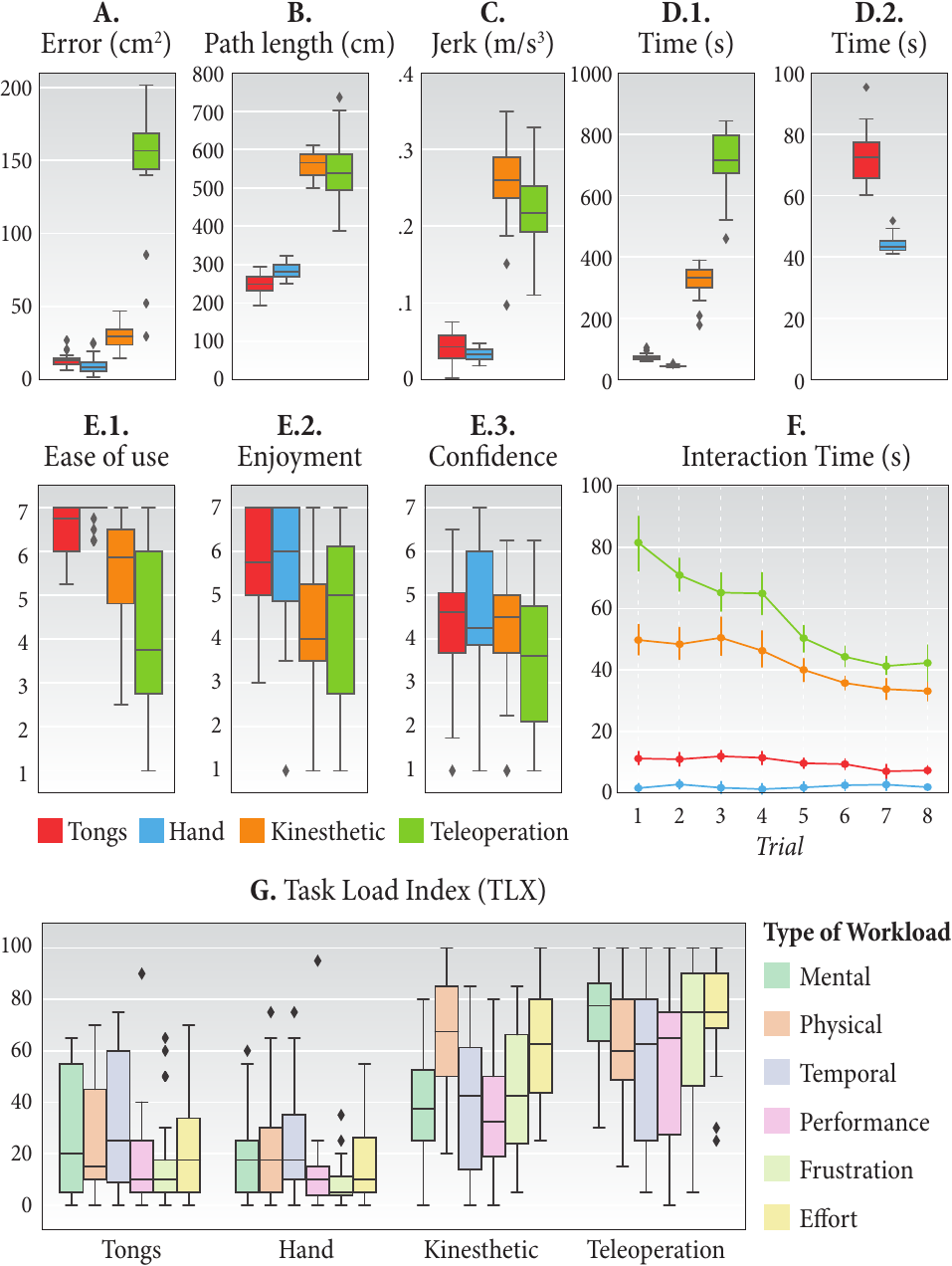}
  \caption{Boxplots for (A) accuracy,  (B) path length, (C) jerk, (D) interaction time, and (E) subjective measures of ease of use, enjoyment and confidence for the four methods of demonstration. (F) Point plot of interaction time during the training task. Vertical bars show standard deviation. (G) Boxplots of the perceived workload on each subscale of the NASA Task Load Index.}
  \vspace{-3mm}
  \label{fig:alldata}
\end{figure}
}

\newcommand{\figtabone}{
\begin{table}[!b]
\vspace{-4mm}
  \centering
  \caption{Descriptive statistics for all measures.} 
  \includegraphics[width=\linewidth]{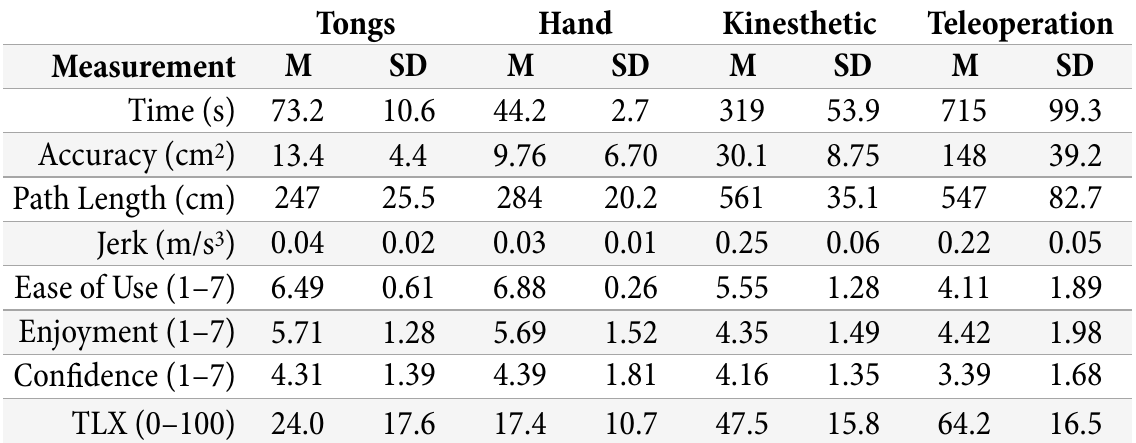}
  \label{tab:des}
  \vspace{-5mm}
\end{table} 
}

\newcommand{\figtabtwo}{
\begin{table}[!b]
  \centering
  \caption{Inferential statistics for all measures.} 
  \includegraphics[width=\linewidth]{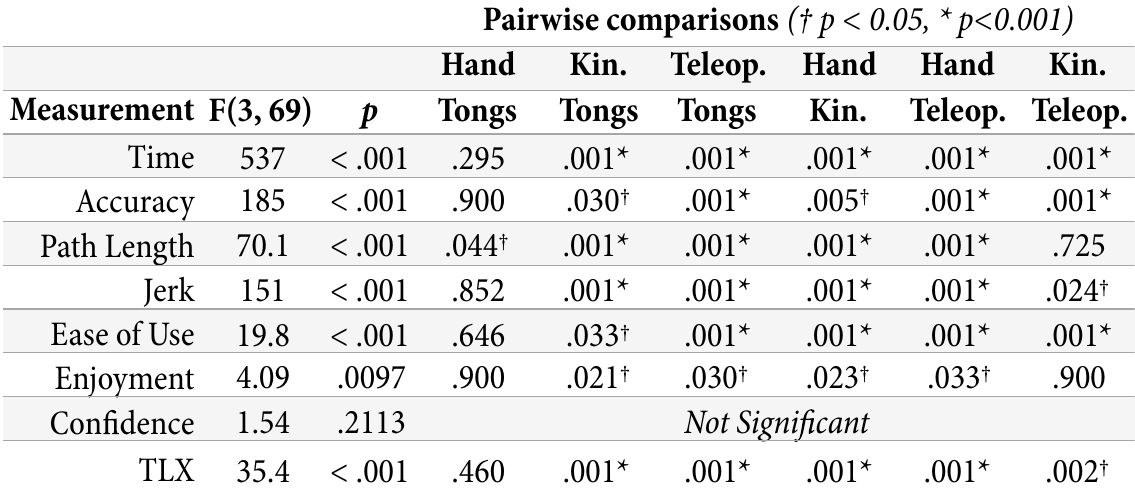}
  \label{tab:inf}
\end{table} 
}

\newcommand{\figusx}{
\begin{figure*}[htb]
  \centering
  \includegraphics[width=\linewidth]{images/user.pdf}
  \caption{Refer to Section \ref{sec:results} for discussion. (A) Boxplots of subjective measures of ease of use, enjoyment and confidence. (B) Point plot of interaction time during the training task. The vertical bars show standard deviation. (C) Boxplots of the perceived workload on each sub-scale of the NASA Task Load Index. }
  \vspace{-1mm}
  \label{fig:usxboxplots}
\end{figure*} 
}
\section{Introduction}

\rp{Demonstrations}{Task \emph{demonstrations}}---examples of users performing actions---are used as input for a range of purposes in robotics, including
programming, guiding, and controlling robots.
Several 
methods 
exist for \rp{providing}{obtaining} these demonstrations, such as \rp{using a joystick}{recording users performing the task with their hands} or
kinesthetically operating the robot by physically moving it. 
Different input methods \rp{}{for demonstrations }have different features, benefits, and drawbacks, making them more or less appropriate for any particular \rp{demonstration scenario}{use case}. 
However, there is no framework to assess this fit 
\rp{towards}{for} 
choosing appropriate methods or determining specifications 
to design new input methods that 
better address the needs of a given \rp{scenario}{use case}.

\rp{Our contribution through this work is twofold. First, we articulate a framework for discussing input methods used for providing demonstrations in the context of hand-scale object manipulation tasks carried out by robot arms with two-finger grippers. Second, to exemplify the utility of this framework, we present and assess the design of a novel input device for such demonstrations, \emph{instrumented tongs}.}{}
\rp{}{In this paper, we present a framework for understanding input methods for obtaining human demonstrations. We use this framework to \rp{assess}{guide} the design of a new input device that addresses a set of goals.
Prior work includes comparisons of the effectiveness of different input methods for specific use cases. For example, Fischer et al. \cite{fischer2016control} compared specific input devices for learning by demonstration, while Rakita et al. \cite{rakita2017mimicry} compared three 
input devices for real-time teleoperation. 
In contrast, we}

\rp{We}{} begin by considering a range of \rp{\textit{usage paradigms} in robotics, such as real-time control or learning from demonstration, that require}{use cases for} human demonstrations\rp{}{in robotics}. From this \rp{list of usage paradigms}{list}, we derive a set of \textit{features} for input methods, each useful in at least some of the\rp{se paradigms}{use cases}. These features provide a \textit{design space} in which we can map existing input methods \rp{in order to characterize them}{}. \rp{In the context of this design space, we also situate}{We introduce} the design of a new input device, instrumented tongs, to support \rp{scenarios}{use cases} where inexperienced users provide demonstrations of object manipulations as input to inform robot execution of these tasks. Such \rp{scenarios}{applications} require input methods that are \rp{}{subjectively }easy for users, provide natural and efficient movements, \rp{support easy instrumentation}{are easy to instrument}, and \rp{achieve good correspondence}{correspond} to robot motions.  
We present a user study that assesses the extent to which instrumented tongs support these goals by comparing it to other commonly used input methods\rp{}{ for demonstrations}. 

\rp{}{Our work makes three main contributions to human-robot interaction (HRI). 
First, we provide a \emph{design space} for demonstration input methods in the form of features that are desirable in robotics applications. 
Second, we present the design of a novel input device for demonstrations, \emph{instrumented tongs}, based on the features identified in our design space.
Finally, our user study provides an \textit{empirical understanding} of the support that our new design and common input methods offer for demonstrations and demonstrators.
}

\section{Related Work}\label{sec:related}
\rp{We draw on prior work to develop two levels of abstraction that serve as the basis for our framework: paradigms for the use of human-to-robot demonstrations \textit{(usage paradigms)}, and desirable qualities of demonstrations and the methods to input them \textit{(features)}. We then use these abstractions to characterize existing and new input methods.}{}

\rp{Prior work has compared specific input devices for learning from demonstration \cite{fischer2016control, kramberger2014lfd, tykal2016kines} and for real-time control \cite{rakita2017mimicry,elton2011hydraulic,kent2017teleop}. We look for other broad categories in the robotics literature that utilize demonstrations and list them as \textit{usage paradigms} in Section \ref{sec:applications}.}{} 

Prior research in HRI has considered either multiple implementations of a single input method \rp{such as kinesthetic guidance or teleoperation}{} (e.g., \cite{akgun2012keyframe, wrede2013user, elton2011hydraulic, kent2017teleop, tykal2016kines, rakita2017mimicry}) or compared multiple input methods \rp{including GUI, space mouse, and haptic interface (e.g.,}{} \cite{fischer2016control, kurenkov2015gui, muxfeldt2014kinesthetic, akgun2012teleop, kramberger2014lfd, muxfeldt2017assembly}\rp{)}{for a single use case (e.g., programming) or task (e.g., assembly)}. At a high level, these studies indicate that input methods affect the experience and quality of the demonstrations provided by users. We build on this body of work to identify \rp{desirable \textit{features}, as listed in Section \ref{sec:features}.}{features desirable across a range of use cases and input methods.}

Many studies provide evidence of features and tradeoffs of specific input methods\rp{, which we take into consideration when generating the design space}{}. For example, Muxfeldt et al. \cite{muxfeldt2014kinesthetic} suggest that extracting human assembly strategies from kinesthetic teaching leads to distorted results as compared to more natural input methods.
Fischer et al. \cite{fischer2016control} discuss issues such as performance, physical exertion, and naturalness across input devices. \rp{}{In the next section, we summarize the issues seen in prior work as features across input methods.}\rp{Prior work that identifies drawbacks of input methods also contributes to our understanding of useful features of input methods. For example, Akgun et al. \cite{akgun2012keyframe} show that kinesthetic demonstrations suffer from extraneous movement, and Pervez et al. \cite{pervez2017teleop} highlight the problems associated with inconsistent demonstrations from teleoperation.}{Other work identifies drawbacks of input methods and provide methods to mitigate them. For example, 
Akgun et al. \cite{akgun2012keyframe} mitigate extraneous movement from kinesthetic demonstrations by using keyframes.
Pervez et al. \cite{pervez2017teleop} address inconsistent demonstrations from teleoperation with learning algorithms. In contrast, we provide ways to choose input devices that may not suffer from these issues.}

\rp{We also draw insights for generating our framework from work in human-computer interaction that involves systematizing human-machine input devices \cite{card1990designspace, card1991designspace}, and}{
Finally, prior work involves} designing specialized devices to address input needs, most notably to support teleoperation \rp{\cite{fischer1990teleop, massie94thephantom, kessler1995cyberglove}}{}.

\rp{}{Our work is a novel approach in HRI for characterizing devices used to provide demonstrations. It considers a broad set of design goals to outline a design space and to guide the design of novel input methods.}

\section{\rp{A Framework for Characterizing \\Input Methods}{DESIGN SPACE}} \label{sec:space}
In this section, we present a \rp{framework}{design space} for \rp{characterizing}{} input methods used for demonstrations in robotics (\mbox{Figure \ref{fig:framework}}). \rp{
The framework emerged from our experiences designing new input methods.
We realized the need for a more formal framework to provide a tool for 
identifying requirements and understanding existing methods.
To create the framework, as described in Section \ref{sec:related}, we distilled insights from prior work into the abstractions of \textit{usage paradigms} and \textit{features} that allow us to characterize commonly used demonstration methods. 
As a starting point, we focus on scenarios 
where demonstrations are provided to a robot arm with a two-finger gripper that performs hand-scale object manipulation tasks. Such scenarios are frequent in current robotics applications.

}{We identify a number of use cases for 
demonstrations and the features necessary to support them. 
We then use the design space (that is made up of these features) to assess the ability of commonly used demonstration methods to serve these use cases and features.}


\subsection{\rp{Usage paradigms}{Use Cases}} \label{sec:applications}
\rp{We identify distinct categories in robotics that use demonstrations and have different needs from them. We refer to the categories as \textit{usage paradigms}.}{We categorize use cases with needs for demonstrations.}



\newcommand{\usefeature}[1]{#1}
\newcommand{\numlist}[1]{\subsubsection{#1}}

\subsubsection{Direct Replay} In direct replay, the robot executes the demonstration as it was recorded, with minimal processing.
It is commonly used to program industrial robots and to 
assess demonstrations in other \rp{usage paradigms (e.g., \cite{akgun2012keyframe})}{use cases \cite{akgun2012keyframe}}.
\rp{}{Direct replay requires demonstrations to serve as effective robot actions.}

\subsubsection{Real-time Control} Demonstrations can enable an operator to control a robot in real time. 
\rp{This}{Real-time control} is also important for Wizard-of-Oz prototyping \cite{riek12woz}.
This usage paradigm requires robot execution that is faithful to the demonstration and that retains the expressiveness of the demonstration.


\subsubsection{One-shot Learning} A single demonstration can be used to inform the generation of actions\rp{}{for a variety of situations (e.g.,\cite{duan2017oneshot})}. This \rp{usage paradigm}{use case} differs from direct replay because the robot will not execute the exact demonstration. Unlike more traditional, multi-demonstration learning, one-shot learning relies heavily on the properties of that single demonstration. 


\subsubsection{Learning from Demonstration (LfD)} We use LfD to refer specifically to \rp{scenarios}{use cases} that generalize from \textit{multiple} demonstrations. Here, demonstrations are not directly executed by the robot but rather combined to create robot movements. 


\subsubsection{Incremental Learning from Demonstration (ILfD)} ILfD refers to a special case where the demonstrations are provided during the learning process. 
ILfD has similar needs to LfD, but may require demonstrations that are responsive to problems identified by the user (e.g., \cite{Bajcsy2018a}) or partial task demonstrations.

\subsubsection{Apprenticeship Learning (AL)} We use this term for \rp{scenarios}{use cases} where demonstrations are only part of the input into the learning algorithm. For example, demonstrations can guide reinforcement learning \cite{kober13rl} either for reward shaping or as starting points for guided exploration.
AL has different needs because of the less direct use of the demonstrations.


\subsubsection{Design} Demonstrations may be used for intermediate steps toward design goals such as identifying reusable primitives for preprogramming robots (e.g., \cite{kabir2016primitives}) or understanding human performance to inform robotics tasks (e.g., \cite{konidaris12cst,Huang2018}).
\rp{This is a distinct usage paradigm because the demonstrations}{These use cases differ from other categories in that they} are not translated into robot motions but rather used to gain insight on demonstrator behavior or performance.

\figframework
\subsection{Features} \label{sec:features}
Each \rp{usage paradigm}{use case above} suggests features that it needs from demonstrations and the methods to input them.
The union of these features provides a list of desirable qualities for input methods. 
\rp{}{Different use cases may have different priorities for these features.}
The features are often interconnected (i.e., are not orthogonal). For example, usability concerns may lead to demonstration quality effects. However, the extended set of features is valuable because there may be multiple ways to achieve a particular feature, and a feature might serve different goals. We organize our discussion by groups of feature types, and label specific features with numbers for later reference. 

\featgroup{Qualities of the Resulting Demonstrations} 
Input devices influence the qualities of the resulting demonstrations by providing affordances that encourage particular properties.
Even if it is possible to add properties to a demonstration \emph{post-hoc,} such improvements come at a cost of reducing how much information is kept from the original demonstration.

\subsubsection{Efficient} \label{sec:feat:efficient}
Many \rp{scenarios}{use cases} \rp{require}{benefit from} demonstrations that have desirable quantitative measurements, such as being short in distance\rp{ or}{, short in} time, low in energy consumption, or low in jerk. Efficient demonstrations may lead to more effective robot executions when replayed or built into learned behaviors.

\subsubsection{Subjective performance} \label{sec:feat:subjperf}
\rp{Many scenarios}{Use cases} benefit from demonstrations that have properties that are difficult to measure directly. For example, they may seek demonstrations that are \emph{natural} (performed in a manner that a person would normally perform the task), or \emph{recognizable} (the demonstrator should see that the movements are similar to what they provided). 
Naturalness may be important if the robot's performance needs to be interpretable by collaborators \cite{vogt17human}, to train robots to make human-like movements (e.g., for communication \cite{bremner2016gestures})\rp{}{\cite{akgun2012keyframe}}, or if the goal is to understand how people perform tasks.

\subsubsection{Desired demonstrations}
\rp{Usage paradigms}{Use cases} often need certain types of demonstrations. For example, they may require \emph{successful} executions of tasks (although failed demonstrations can also be learned from \cite{Grollman2012}). In some cases, meaningful \emph{variance} in executions is useful to understand how to generalize beyond the provided demonstrations
(e.g., \cite{calinon13extra}, \cite{lee2015force}).
In contrast, extraneous variance such as differences between executions because of false starts, inflections, mistakes, or inaccuracies, are less often useful, although some analysis techniques might exploit such information (e.g., \cite{rozo16gmm}).

\subsubsection{Amenable to analysis}
Different downstream \rp{data processing algorithms}{applications} may require different \rp{input}{} data properties. For example, some \rp{scenarios}{use cases} need continuous trajectories while others employ only keyframes. Different input devices may have different affordances for different data types, such as being more or less easy to create continuous movements \cite{akgun2012keyframe} or providing the ability to capture meta-data to inform analyses such as segmentation and keyframe determination.

\featgroup{Qualities of the Demonstrator Experience}
The input device affects the experience of giving a demonstration. 
\rp{}{While ``naturalness'' is often sought as a goal for input devices, we instead focus on the needs of use cases for which 
may be achieved by multiple strategies.}
The importance of the demonstrator experience varies, e.g., scenarios with casual users may place more emphasis on the subjective experience than methods designed for experts performing critical tasks.

\subsubsection{Affords quality demonstrations}
As described above, the input device may facilitate a person to give demonstrations  \emph{naturally} (as they would in the real-world task), \emph{confidently} (to know that they are achieving their goals), and \emph{expressively} (to achieve other demonstration goals such as speed).

\subsubsection{Facile} The input method should place low mental and physical demands on the demonstrator
beyond the task itself.
For example, expecting a demonstrator to lift a heavy robot or attend to collisions of the robot's joints in the environment may distract them from performing a challenging task.
\rp{}{Scenarios involving repetition or complex tasks are particularly likely to benefit from having limited user demands.}

\subsubsection{Easy to learn} Experience in using a device often leads to better performance. However, some devices allow demonstrators to achieve good performance with less training.

\subsubsection{Preference} Many scenarios value input methods that demonstrators like.
For example, in casual user scenarios, methods that are novel or fun may encourage users to try an application. Conversely, casual users may stop using things they dislike when they aren't being compelled.

\subsubsection{Feedback} The input method must provide the demonstrator sufficient feedback to perform the actions they intend. Visual, haptic, tactile, and even auditory cues can be important to helping a demonstrator perform a task.

\featgroup{Correspondence with Robot} In many \rp{usage paradigms}{use cases}, demonstrations are used to create robot actions, requiring them to be relevant to how the robot will perform the actions. This correspondence can be challenging when the demonstrator has different abilities than the robot, such as the different kinematic and dynamic properties of human and robot arms. \rp{}{Differences can be particularly acute for manipulation tasks as the human hand has dozens of actuators and thousands of 
sensors \cite{Controzzi2014,Balasubramanian2014}. Dexterous robot hands are possible \cite{Xiong2016}, however simple designs are likely to remain a common approach \cite{Odhner2014,Odhner2015}.}

We distinguish correspondence for demonstrations into two types: plausible and feasible. \emph{Plausible} demonstrations match the capabilities of the target robot, whereas \emph{feasible} demonstrations can be executed by the robot. For example, in a pick-and-place demonstration, an implausible demonstration may involve grasping the object with multiple points of contact on a single finger, using dexterous in-hand manipulations to orient the object, or using fine haptic sensing to determine where on the object to grasp. Even with a plausible demonstration, e.g., performing the above task using a pinch grasp and a wrist movement to correctly orient it, the robot may not be able to feasibly follow the paths of the fingers due to kinematic constraints, collisions, or singularities.

\rp{}{An input device may enforce correspondence, e.g., by requiring the user to move the physical robot through the motions. Similarly, a device may merely encourage correspondence, e.g., asking a user to only use pinch grasps when giving a free-hand demonstration.}

\subsubsection{Plausibility}
It is often valuable that the basic strategy of performing a task is a plausible one for the robot to use. While it may be possible to transform an implausible demonstration into a plausible one, these transformations are necessarily complex (because they involve changing strategy). Even if the transformation is possible, it may change the demonstration enough\rp{}{that} to disturb other properties such as recognizability.

\subsubsection{Feasibility}
Feasibility ensures that a specific demonstration can be executed on the robot. \rp{}{Feasible demonstrations are required for a robot to perform a demonstration, e.g., in direct replay or real-time control. In other use cases, demonstrations are processed to produce trajectories. For example, in LfD, the motions executed by the robot are generated from models created from the demonstrations. However, in such scenarios, replay may be used to allow users to validate demonstrations (e.g., \cite{akgun2012keyframe}).}
Feasible demonstrations are enforced by methods that involve physically moving the robot. However, given a plausible end-effector path, feasible trajectories can be generated through optimization (e.g., \cite{Luo2012,Rakita2018rss}), and near-feasible trajectories can be made feasible through trajectory optimization (e.g., \cite{Luo2012,Luo2015,Dragan2011,Kalakrishnan2011,Zucker2013}).
Such computational methods for creating feasible motions are necessary in scenarios that generalize demonstrations because \rp{even}{the executed trajectories are not demonstrations. Even} if the demonstrations are feasible, the trajectories generated by a learning method may not be.

\featgroup{Practical Qualities} Practical concerns make up the final category of features, that is if the device can be built and deployed in an effective manner. We focus on one specific aspect of practicality 
that concerns input \rp{devices}{modality}.

\subsubsection{Instrumentable}
\rp{All usage paradigms}{Most cases} require measurements of \rp{}{the} demonstrations using appropriate sensors, including positions, orientations, and applied forces and torques. Different \rp{scanerios}{use cases} may require different types or fidelity of measurements. For example, \rp{understanding}{a use case to understand} human task strategies may only need coarse timing data, while a method to infer constraints from a demonstration may require precise position, orientation, force, and torque measurements \cite{guru2018icra}. Different input devices 
afford different instrumentation.
For example, positioning a robot provides precise kinematic information whereas tracking the position of the demonstrator's fingers may be challenging.
\subsection{Characterizing Input Methods} 
\label{sec:inputs}
\rp{In this section we first discuss two points about usage paradigms and features that are applicable to the process of characterizing input methods. 

\textit{How do usage paradigms relate to features?} Different usage paradigms may require different features from demonstrations. Within this discussion, we look at binary weightings of these features, that is whether a feature is prioritized or not. For instance, direct replay and real-time control require robot execution to be faithful to the demonstration and retain the expressiveness of the demonstration (e.g., \cite{haage2017teaching,lin2015remote}). These paradigms would benefit from efficient (Feature 1), recognizable (2), plausible (10) and feasible (11) demonstrations. In addition, specific scenarios will require the presence of other features. Scenarios involving repetition or complex tasks are likely to benefit from having input methods that are facile (6), for instance to provide multiple knot tying demonstrations for LfD \cite{lee2015force}. An easy to learn (7) input method may allow non-expert users to confidently (5) provide demonstrations for one-shot learning (e.g., \cite{orendt2016oneshot}). In Bajcsy et al. \cite{Bajcsy2018a}, demonstrations are required in the form of corrections as input for incremental LfD. Some of the prioritized features are sporadic demonstrations (3), plausibility (10), feasibility (11), recognizability (2), capturing corrections (12) and feedback (9). It is also possible that certain features are intentionally de-prioritized, for instance failed demonstrations are encouraged in Grollman and Billard \cite{Grollman2012} for improving LfD algorithms.}{}

\rp{\textit{How can features be used in the process of characterization of input methods?} Each feature can be evaluated using different measures. For instance, in Fischer et al. \cite{fischer2016control}, length of demonstrated paths and interaction times are used to measure efficiency (Feature 1). Kramberger \cite{kramberger2014lfd} used standard deviation of the learned trajectories to measure repeatability (3). Muxfeldt et al. \cite{muxfeldt2014kinesthetic} and Rakita et al. \cite{rakita2017mimicry} use decreasing interaction time of repeated trials as an indication of ease of learning (7). A detailed discussion of the measurements for each feature is beyond the scope of this paper. However, Section \ref{sec:study} provides an example of a broad set of measures that may be used to measure the quality of demonstration and demonstrator experience.}{}


\rp{Below, we theorize from prior studies referred to in Section \ref{sec:related} about three input methods commonly used to provide demonstrations across a range of robotics applications. Our empirical evaluation of these input methods is part of the user study described in Section \ref{sec:study}.}{}

\subsubsection{Kinesthetic \rp{Guidance (KG)}{Demonstrations (KD)}}
In KG, the demonstrator physically moves the robot.
Such an input method requires that the demonstrator has access to the robot system and that the robot can be ``back-driven'' to be posed manually. Because robots can typically sense their configurations, KG inputs provide precise kinematic measurements and, in some cases, force information (Feature 12). KG provides plausible (10) and feasible (11) demonstrations, as the use of the robot to create the movements ensures that it is capable of performing them. 

User experience with KG depends on the implementation; some robots are easier to move (e.g., due to better gravity compensation) while others have kinematic properties that are hard to understand. At a high level, KG requires the user to consider how the robot moves during the demonstration, particularly to create feasible movements. The need to physically move the robot can negatively affect user preference (8), ease of learning (7), facileness (6), and demonstration qualities (5). While KG generally offers good visual feedback, it does not provide users with a complete haptic sense of the robot's performance (9). Although prior studies \cite{wrede2013user, muxfeldt2014kinesthetic} have shown user preference toward KG over traditional input methods (e.g., joystick or teach pendants), these studies do not include more natural input methods as we will in Section~\ref{sec:assessment}. 

The qualities of movements obtained with KG are hard to assess \emph{a priori.} We expect the unnatural mode of input and the need to consider the robot's kinematics to be a challenge in creating demonstrations that are efficient (1), natural (2), or of the desired type (3), particularly for inexperienced users. Motion artifacts, e.g., irrelevant starts and stops to change grip on the robot, can make analysis challenging (4). 

\subsubsection{Free-Hand Manipulation (FHM)}
In FHM, users perform demonstrations with their hand. Unlike in KG, obtaining precise kinematic measurements \rp{and force information}{} of these demonstrations is difficult (12). Instrumented gloves are also expensive and uncommon.
Robustly tracking fingers from video or optical data is challenging because of occlusion.
\rp{}{Measuring force information is also challenging.} Because the hand is so capable, the resulting demonstrations may not be plausible or feasible for robot execution (10, 11).

Our inherent skill and lifelong experience at using our hands to perform actions leads to high quality demonstrations (5) with little extra effort (6) and no learning (7). We expect people to also show higher preference (8) toward using this inherent ability. The hands provide considerable amount of sensing (9). 
People's inherent skill in using their hands also ensures that the movements will be efficient (1), natural for them (2), and controllable to achieve desired goals (3). 

\subsubsection{Teleoperation (TO)}
In TO, the demonstrator is physically separated from the space where the demonstration is performed. TO often involves real-time remote control of a robot or its virtual representation for real-time or future use. Unlike in FHM, the movements are used to control a robot, and therefore perform tasks indirectly. The specific input device used to control the robot can vary; see Rakita et al. \cite{rakita2017mimicry} for a comparison of various options. 

Feedback (9) is a critical issue in TO, as the operator is unable to directly observe the action. Prior work has explored techniques for visual \cite{rakita2018camera} and haptic feedback \cite{haptic-survey}. Similar to KG, TO is naturally instrumented to record kinematic information (12) and provides plausible (10) and feasible (11) movements (assuming that the actual robot is used). Prior work has explored natural interfaces for teleoperation control and compared it to other standard interfaces \cite{rakita2017mimicry}. While this interface has been shown to have usability and performance advantages relative to other TO input devices \cite{rakita2017mimicry}, its performance relative to other non-TO demonstration approaches is unknown. 




\section{Design \rp{of Instrumented Tongs}{}} 

\rp{}{This section provides motivation for and describes the design of a new input device for demonstrations for robotics applications.}
Our vision is to enable inexperienced users to teach robots to perform physical manipulation tasks and this
highlights the need to satisfy the following design goals.

\subsection{Design Goals}
\label{sec:goals}

\subsubsection{Obtaining High-Quality Demonstrations}
Efficiency is valuable (1) as we seek to execute demonstrations (or programs derived from them) on the robot. Natural movements (2) will enable the demonstrator to easily verify robot actions. Most \rp{scenarios}{methods} we consider require successful demonstrations, and some benefit from meaningful variance (3).

\subsubsection{Improving User Experience} Because we are interested in supporting inexperienced users, we are concerned with the qualitative experience of providing demonstrations. We want to enable inexperienced users to provide quality demonstrations of potentially complex tasks. Our input device must be facile (6), easily to learn (7), and liked (8). \rp{Tactile}{Some degree of tactile} feedback is valuable to complete complex manipulation tasks effectively (9).

\subsubsection{Trajectory Quality} Because demonstrations may be executed on the robot, we require plausibility (10), and to a lesser degree, feasibility (11). We can use optimization to create feasible trajectories from end-effector paths.

\subsubsection{Precise Measurement} Many robot control algorithms require precise measurement of position and orientation as well as applied forces and torques during the demonstration (12). 

Unfortunately, existing methods do not serve this set of desired features. Free-hand manipulation offers the desired user experience and movement qualities, but is too difficult to instrument and does not encourage plausible demonstrations. Kinesthetic demonstrations offer easy instrumentation and enforce plausibility. However, we were concerned that its unnaturalness would not provide the desired user experience and would lead to poor quality movements in the demonstration.


\subsection{Design Details}
\label{sec:design}
Our design is inspired by kitchen tongs. We observe that tongs restrict manipulation to a clumsy form of pinch grasp that is simpler than most robot grippers. Despite this limited capability, people use tongs adeptly to perform a wide range of manipulations (e.g., patrons at a salad bar or a parent pulling a splinter from a child with tweezers). Therefore, tongs serve as a metaphor for the limitations of a robot gripper. 

\rp{The design details in this section}{These details} are specific to the prototype shown in Figure \ref{fig:tongsdesign} and used in the experiments of Section \ref{sec:assessment}.
\rp{}{Our initial design used the kitchen tongs metaphor literally; we adapted standard kitchen tongs by attaching pressure sensors in the tips and an inertial measurement unit (IMU) near the hinge. Kinematic data (position and orientation) was approximated using  algorithms from VR tracking applications \cite{LaValle2014}. The prototype showed the need for an improved design that rigidly mounted high quality sensors.} The specific details may be adapted based on the sensors available and the size of objects to be manipulated. Full construction plans, including shape files for 3D printing and assembly suggestions, are available under a non-restrictive open-source license.\footnote{https://github.com/uwgraphics/HRI2019Tongs}

\subsubsection{Basic Construction} 
The instrumented tongs consist of two rigid, 3D printed arms made from PLA plastic. 
A hinge consisting of a steel pin with plastic bushings holds the arms together. 
The tongs are held open with a flat \emph{spring steel} spring.
On the grasping side of the tongs, two inset mounting surfaces enable attachment of force-torque sensors. Multiple prongs provide locations to attach motion capture markers. 

\subsubsection{Size and Mechanism} Our tongs are sized to be similar to\rp{}{the} kitchen tongs.\rp{}{from our initial prototype.} We felt this size fit comfortably in the hand and provided the ability to manipulate \rp{hand-scale objects}{objects of a useful range of sizes}. 
Our tongs retain the single hinge design from kitchen tongs, which is a simpler mechanism than the parallel jaw, compliant, or under-actuated designs commonly seen in robot grippers. The simple hinge design has advantages\rp{;}{ in that} it is easy to design and build, requires few parts, is familiar to most users (as it mimics common tools such as pliers, tweezers and kitchen tongs), and provides for direct haptic feedback. 

\subsubsection{Gripper Surface} 
The hinged geometry described earlier makes the jaws parallel in only one position. Rigid objects contact the gripper at two points. To mitigate this issue, we build compliance into the tongs by padding the contact surfaces with dense foam to provide enough contact area to manipulate rigid objects such as blocks in Section \ref{sec:assessment}, yet firm enough to provide good tactile cues.
A plastic plate is attached directly to the tool side of the force-torque sensors to provide a mounting surface for softer padding. The plastic plate is wedge shaped to accommodate a relatively parallel grasp of large objects. 
The padding consists of a low-density thick foam layer (\texttildelow 10mm) followed by a high density thin foam layer (1mm).
A final layer of 3M Gripping material (3M TB400) was used to provide sufficient friction to hold objects in place. 

\subsubsection{Sensors}\label{sec:sensors}
The current version of the tongs uses an Optitrack motion capture system which employs reflective markers that are rigidly attached to the tongs and cameras to record the motion of these markers. We track both arms as independent rigid bodies. 
A pair of ATI Mini40 force-torque sensors, one in each arm, provide accurate measurement of the applied forces and torques when grasping objects. 
The current sensors provide precise measurement of forces and torques applied by the tongs to a manipulated object. 
While force-torque sensors are highly accurate instruments (which directly reflect on cost and durability), many applications may not need the accuracy or may require a different kind of measurement (e.g., temperature, vibrations).


\rp{}{A first iteration used small force sensors and an HTC Vive controller to track position and orientation. A second iteration used better sensors as discussed \rp{in Section \ref{sec:details}}{below}.}


\label{sec:details}
\rp{}{In this section we describe the specific details of our prototype instrumented tongs.
The main design elements of the tongs are discussed in the previous section: the choice of a simple hinge design, sized similar to kitchen tongs, and shaped to rigidly attach sensors.}



\figdesign
\section{Assessment}
\label{sec:assessment}

This section describes a study to validate that instrumented tongs meet the design goals listed in Section \ref{sec:goals} by comparing the tongs against three methods commonly used in human-to-robot demonstrations discussed in Section \ref{sec:inputs}. The study aimed to determine whether the tongs could provide the demonstration quality and user experience of free-hand manipulation with the mapping and practicality benefits of kinesthetic demonstrations. We developed a set of object-manipulation tasks and asked participants to demonstrate 
these tasks with 
four input methods: free-hand manipulation, kinesthetic guidance, and teleoperation as described in Section \ref{sec:inputs}; and instrumented tongs as described in Section \ref{sec:design}.



 \subsection{Study Design} \label{sec:study}


\featgroup{Experimental Design} We designed a $4 \times 1$ within-participants experiment in which novice participants provided demonstrations of three tasks with all four input methods\textemdash \textit{Tongs}, \textit{Hand}, \textit{Kinesthetic}, and \textit{Teleoperation}\textemdash in a counterbalanced order. Movements during the demonstration were recorded using an Optitrack motion capture system that tracked reflective markers. 
Synchronous video was captured by a ceiling-mounted \rp{Logitech C920 Pro}{} webcam. Participants filled out questionnaires after using each input method. 

\featgroup{Participants} We recruited 24 participants (13 male, 11 female) from the University of Wisconsin--Madison campus between the ages of 21 and 27 ($M = 23.75$, $SD = 1.54$). Participants reported some familiarity with robots ($M = 3.17$, $SD = 1.39$, measured on a 7-point scale). Five participants reported an interaction with a robot in prior robotics research studies. The study took 90\textendash120 minutes, and all participants received USD 20 as compensation.

\featgroup{Methods of Demonstration} 
We used \rp{representative}{the following specific} implementations of the methods discussed in Section \ref{sec:inputs}\rp{ that allow us to assess general properties of each method}{}.

\subsubsection*{Hand} Participants wore a glove that did not interfere with the dominant hand's functional dexterity. All participants wore the glove on the right hand (23 participants were right-handed, and one participant was ambidextrous). Motion-capture markers were attached to the back of the glove to track hand movement but not finger movement, allowing us to assess the amount of arm movement but not the details of manipulation.

\subsubsection*{Kinesthetic} Participants used a UR5 robot arm from Universal Robots equipped with a Robotiq 85 two-finger gripper. The robot was placed in ``freedrive'' mode that allowed for it to be moved freely with gravity compensation. A separate button was provided on a wireless remote to open and close the gripper. While the joint angles of the robot can be recorded, we tracked the end-effector position and orientation using the same motion capture markers as with other devices.

\subsubsection*{Teleoperation} Participants used the system described by Rakita et al. \cite{rakita2017mimicry} to teleoperate the UR5 robot. The participants held an HTC Vive controller that tracked the position and orientation of their hand, which the system mapped to the robot end-effector in real time, causing the robot to mimic the natural arm motions of the participant. A button on the controller operated the robot gripper.
To ensure safety, participants stood outside the working radius (850 mm) of the robot. 

\subsubsection*{Instrumented Tongs} Participants used their dominant hand to provide demonstrations, and motion-capture markers on the tongs were used to record them. 

\figteaser

\featgroup{Setup, Tasks, \& Procedure} Tasks involved manipulating $5\times2.5\times1.5$-inch ($L\times W\times H$) blocks in a workspace across various locations identified by the letters P, Q, R, and T (Figure~\ref{fig:teaser}). Participants performed three tasks in a fixed order:

\subsubsection{Training task}  For each trial, the experimenter places a foam block in one of the starting locations (P, Q, or R). The participant picks this block up and places it on the target (T). The task consists of eight trials.


\subsubsection{Foam block stacking} Foam blocks are placed at positions P, Q, and R. The participant picks each one up and places them at position T, creating a stack. The third block is upside down and must be flipped before being placed.

\subsubsection{Lego block stacking} This task is similar to \textit{foam block stacking}, except that large Lego blocks are used in place of foam blocks. These blocks must be snapped together, requiring precise alignment and appropriate forces.


The procedure was administered under a protocol reviewed and approved by the University of Wisconsin--Madison Education and Social/Behavioral Science Institutional Review Board (IRB). Following informed consent, participants watched a video outlining the experiment goals and then used each input method in the order assigned to them. For each method, participants first watched a training video, then performed the three tasks, and finally filled out two online questionnaires.


\featgroup{Measurement and Analysis} Here, we describe\rp{}{ a set of objective and subjective} measures to assess support for our design \rp{goals}{requirements}. \rp{Interaction time, accuracy, path length, and jerk are captured for the block-stacking tasks, learning effect for the training task, and user preferences and workload at the end of all tasks}{All metrics except for \emph{Learning Effects} are captured for both stacking tasks but not the training task}.


\subsubsection*{Interaction Time} We measured the total amount of time taken to complete the foam and Lego block-stacking tasks. 

\subsubsection*{Accuracy} We measured the average accuracy of the final placement of the blocks \rp{at}{with respect to} the targets by calculating the area of the target rectangle that remains exposed after the blocks were placed over it. 
Both position and rotation error were captured by this metric. If a block was not placed at the final position, the error was capped at the total area of the target position. 

\subsubsection*{Path length} Total distance traveled by the hand, averaged across the two tasks, served as a measure of task efficiency.

\subsubsection*{Jerk} We used average jerk as a measure of extraneous movement. Jerk was calculated as the time derivative of the acceleration of the captured trajectory. 
Points missing from the data (e.g., from motion capture occlusions) were omitted.

\subsubsection*{Learning effect} Change in the time taken over multiple trials served as a measure of learning.


\subsubsection*{User preferences} We administered a questionnaire based on prior research on measuring user preferences \cite{davis1989perceived}. The questionnaire included three scales, each with four items measured on a seven-point rating scale (1 = strongly disagree; 7 = strongly agree), to measure ease of use of the input method (Cronbach's $\alpha$ = 0.94), enjoyment (Cronbach's $\alpha$ = 0.89), and confidence that the robot will be able to learn from the demonstration (Cronbach's $\alpha$ = 0.94). 

\subsubsection*{Workload} We used the NASA Task Load Index (TLX) \cite{hart1988development} to assess user's perceived workload. The total workload is divided into six subscales---mental demand, physical demand, temporal demand, performance, effort, and frustration---that are measured on a 100-point rating scale, where higher ratings indicate more workload. The scores are aggregated with equal weighting to calculate the TLX. 

We analyzed data from all measures using one-way repeated-measures analyses of variance (ANOVA) to determine whether or not the input method had a significant effect. If the ANOVA test showed significant differences, we used Tukey's HSD test in order to determine where the differences lied while accounting for multiple comparisons.

\subsection{Results and Discussion} \label{sec:results}

This section is organized according to the design goals set in Section \ref{sec:goals}. The descriptive and inferential statistics for the results described below are shown in Tables \ref{tab:des} and \ref{tab:inf}. 

\figall

\featgroup{Quality} 
Quantitative motion quality metrics are summarized in Figure \ref{fig:alldata}.A\textendash D. Across all metrics\textemdash accuracy, path length, smoothness, and interaction time\textemdash hand and tongs significantly outperformed teleoperation and kinesthetic demonstrations. More surprisingly, tongs were competitive with hand for all metrics. Differences were not significant, except a shorter average path length with tongs than with hand. Because the significance tests consider variance across the whole experiment, they smooth over the detail shown in Figure \ref{fig:alldata}.D.2, which suggests that tongs are consistently slightly slower than hand. We discuss this observation under \textit{plausibility} below. This analysis suggests that tongs meet the motion quality goals as they are competitive with the best of the prior methods. Although we did not directly assess its naturalness, its similarity to hand suggests tongs to be a natural input method.



\featgroup{User Experience} We found that the goals for a better experience for the demonstrator were met: the tongs allowed users to easily provide expressive demonstrations without prior experience. These results are shown in Figure \ref{fig:alldata}.E\textendash G.
Tongs and hand provided significantly higher ease of use and enjoyment, and lower workload than teleoperation or kinesthetic guidance. However, there were no significant differences between hand and tongs.
We found no significant differences among the methods in the participants' perceived confidence in the robot's ability to perform a similar task in the future. 

Perceived workload, as measured by the TLX scores, followed the inverse trend as perceived ease of use, indicating an association between ease of use of the input method and demonstrator workload. 
Participants found kinesthetic guidance physically strenuous, consistent with the higher jerk and path length observed with this method. Participants found teleoperation control mentally demanding, perhaps because it required them to constantly update their understanding of the correspondence between their movements and the robot. 

Interaction time during the training task shows that participants were already adept at using their hands and tongs. However, they adapted and improved performance across multiple trials for other methods. 
In our experimental trials, we observed participants to rely on haptic feedback during demonstrations using \rp{hands or tongs}{the instrumented tongs as well as manual demonstrations}. Often, \rp{participants fumbled}{these demonstrations resulted in participants fumbling} during the assembly of blocks and \rp{depended additionally}{depending} on visual feedback for better alignment.

\featgroup{Correspondence} 
Kinesthetic and teleoperation both provide demonstrations that have a direct correspondence with the robot. 
While mapping position and orientation of hands or tongs to a robot configuration requires complex optimization-based computational approaches (e.g., \cite{Luo2012,Rakita2018rss}), demonstrations provide better \emph{plausibility} with tongs. In our observations, while grasping the block with tongs, participants approached carefully from a direction that allowed a good grip, similar to what a robot would perform. In contrast, during free-hand demonstrations, participants grasped the block in a variety of ways from different directions. Participants also primarily flipped the block using an in-hand manipulation, although some participants used a parallel grip after using other input methods. Using the tongs, the flipping action involved twisting the arm, as it would on the robot. We believe these plausibility differences account for the inefficiency of tongs relative to hand seen in Figure \ref{fig:alldata}.D.2.



\featgroup{Instrumentation} As described in Section \ref{sec:sensors}, our tongs are instrumented to allow for precise measurement of position and orientation as well as force and torque at the contact surface. Tracking the robot during teleoperation or kinesthetic demonstrations also provides precise measurement of kinematic information (gripper position and orientation). The fingers of the robot gripper could have the same sensors installed as the tongs to capture forces and torques. 
On the other hand, our data collection lacked the instrumentation necessary to obtain good kinematic information from free-hand demonstrations. Because of the limitations of our sensing equipment, we were only able to measure the position and orientation of the hand as a rigid body. While these measurements were sufficient for comparing the amount of arm motion between different methods, most other scenarios would require capturing more information about the manipulations. Precisely measuring the positions of the fingertips, even in the relatively uncluttered environment of our experiment, would require sensors beyond what is readily available off the shelf.
\figtabone
\figtabtwo

\featgroup{Summary} Our evaluation showed that the use of tongs as an input method provides many of these features at levels comparable to the best existing input methods. No existing method provides the combination of good demonstration quality, good demonstrator experience, good correspondence with the robot, and good instrumentability, as provided by tongs. While kinesthetic demonstrations have been shown to be effective for robotics applications, our study suggests that they have inferior motion quality and user experience compared to more natural input methods. Similarly, natural teleoperation interfaces that have been shown to be superior to other teleoperation schemes do not provide the advantages of direct natural interfaces.

\section{General Discussion}


Our \rp{framework}{design space } allows us to reason about the potential efficacy of \rp{different input methods, including the}{} instrumented tongs\rp{.}{ for a range of use cases by identifying the features of demonstration input that may be required.} \rp{Because we cannot make claims regarding its comprehensiveness \cite{Kerracher2017}, how well it will generalize to new scenarios is unclear. Additionally, our selection of features is shaped by our focus within this paper on demonstrations of hand-scale manipulations to robot arms with two-finger grippers. Other features may emerge from considering more complex scenarios. For example, the consideration of more types of robot end-effectors would require us to add the ability to retarget the demonstration between end-effectors  to the set of features. Expanding the framework to such scenarios is a promising direction for extension of this work.}{}

\rp{Our study provides an example of an empirical evaluation, including measurement methods and tools, that can be used to understand input methods using this framework. However, it is limited by the specific tasks and device implementations that are considered. We are particularly interested in how our results extend to more complicated tasks in cluttered spaces. New techniques may provide better implementations, such as practical hand tracking in cluttered environments or better mechanisms to make kinesthetic control of robots easier, which may alter the relative merits of their respective methods. The findings from our study suggest that tongs provide a successful solution to a specific set of needs: capturing kinematic and force/torque demonstrations for hand-scale object manipulations. It is unclear how well its use generalizes to a broader range of scenarios. Our current design uses high-precision sensors that are costly and cumbersome and that require cables. The bulkiness of the device limits the tasks and environments in which it can be used. Variations in the design may allow the instrumented tongs to achieve the design goals with different trade-offs for other scenarios. For example, using smaller force sensors (that may be less precise) could provide a smaller device that can support finger-scale manipulations.

The framework introduced in our work provides a tool to characterize input methods for human-to-robot demonstrations. We believe that this approach can be extended to broaden the applicability of the framework. Our new input method, instrumented tongs, has already enabled the development of new techniques (e.g., \cite{guru2018icra, Guru2018corl, Guru2019icra}) through its combination of practicality, usability, and demonstration quality.}{}
\rp{}{
\emph{Limitations of the Design Space:}
While our design space of 12 features provides a tool for understanding how different input devices may fill the needs of different use cases, it may not provide a complete tool for future analyses. Because we cannot make claims regarding the completeness of the space \cite{Kerracher2017}, how well it will generalize to new use cases or to radically different input methods is unclear. 
\emph{Limitations of Instrumented Tongs:}

The tongs provide a successful solution to a specific set of needs: capturing kinematic and force/torque demonstrations for hand-scale object manipulations. It is unclear how well they generalize to a broader range of scenarios. Our current design uses high precision sensors that are costly, cumbersome, and require cables. Choosing sensors with different trade-offs, for example using smaller but less precise force sensors, may provide a less bulky device. The bulkiness of the current design limits the tasks and environments that the device can be used in.  
We have not explored variations of the tongs design. For example, we might use a more complex mechanical design similar to parallel jaw pliers to better approximate a robot gripper, or attach the gripping surfaces more directly to the user's fingers like castanets. As these designs share the strategy of our current design, providing natural but constrained movements, we expect their performance to be similar. 

\emph{Limitations of the User Study:}
While the study suggests the success of tongs at achieving its design goals and provides some evidence for the relative performance of other devices, it is limited. In particular, the study considers specific tasks and device implementations. We are particularly interested in how our results extend to more complicated tasks in cluttered spaces. New techniques may provide better implementations, such as practical hand tracking in cluttered environments or better control mechanisms to make kinesthetic control of robots easier, which may alter the relative merits of their respective methods. Our study also provides no direct evidence that the tongs are useful in a specific application. 
However, we have already seen new techniques (e.g., citations removed for anonymity) enabled by the tongs' combination of practicality, instrumentability, usability, and demonstration quality.}
\section*{Acknowledgment}

We thank Michael Zinn for his guidance in the development of the instrumented tongs, Daniel Rakita for his valuable feedback, and the anonymous reviewers for their suggestions. 

\clearpage

\balance
\bibliographystyle{ieeetr}
\bibliography{references}	

\begin{thebibliography}{10}

\bibitem{fischer2016control}
K.~Fischer, F.~Kirstein, L.~C. Jensen, N.~Krüger, K.~Kukliński, M.~V. aus~der
  Wieschen, and T.~R. Savarimuthu, ``A comparison of types of robot control for
  programming by demonstration,'' in {\em 2016 11th ACM/IEEE International
  Conference on Human-Robot Interaction (HRI)}, pp.~213--220, March 2016.

\bibitem{kramberger2014lfd}
A.~Kramberger, ``A comparison of learning-by-demonstration methods for
  force-based robot skills,'' in {\em 2014 23rd International Conference on
  Robotics in Alpe-Adria-Danube Region (RAAD)}, pp.~1--6, September 2014.

\bibitem{tykal2016kines}
M.~Tykal, A.~Montebelli, and V.~Kyrki, ``Incrementally assisted kinesthetic
  teaching for programming by demonstration,'' in {\em 2016 11th ACM/IEEE
  International Conference on Human-Robot Interaction (HRI)}, pp.~205--212,
  March 2016.

\bibitem{rakita2017mimicry}
D.~Rakita, B.~Mutlu, and M.~Gleicher, ``A motion retargeting method for
  effective mimicry-based teleoperation of robot arms,'' in {\em 2017 12th
  ACM/IEEE International Conference on Human-Robot Interaction (HRI)},
  pp.~361--370, March 2017.

\bibitem{elton2011hydraulic}
M.~D. Elton and W.~J. Book, ``Comparison of human-machine interfaces designed
  for novices teleoperating multi-dof hydraulic manipulators,'' in {\em 2011
  20th IEEE International Symposium on Robot and Human Interactive
  Communication (RO-MAN)}, pp.~395--400, July 2011.

\bibitem{kent2017teleop}
D.~Kent, C.~Saldanha, and S.~Chernova, ``A comparison of remote robot
  teleoperation interfaces for general object manipulation,'' in {\em 2017 12th
  ACM/IEEE International Conference on Human-Robot Interaction (HRI)},
  pp.~371--379, March 2017.

\bibitem{akgun2012keyframe}
B.~Akgun, M.~Cakmak, J.~W. Yoo, and A.~L. Thomaz, ``Trajectories and keyframes
  for kinesthetic teaching: A human-robot interaction perspective,'' in {\em
  2012 7th ACM/IEEE International Conference on Human-Robot Interaction (HRI)},
  pp.~391--398, March 2012.

\bibitem{wrede2013user}
S.~Wrede, C.~Emmerich, R.~Gr\"{u}nberg, A.~Nordmann, A.~Swadzba, and J.~Steil,
  ``A user study on kinesthetic teaching of redundant robots in task and
  configuration space,'' {\em Journal of Human-Robot Interaction}, vol.~2,
  pp.~56--81, February 2013.

\bibitem{kurenkov2015gui}
A.~Kurenkov, B.~Akgun, and A.~L. Thomaz, ``An evaluation of {GUI} and
  kinesthetic teaching methods for constrained-keyframe skills,'' in {\em 2015
  IEEE/RSJ International Conference on Intelligent Robots and Systems (IROS)},
  pp.~3608--3613, September 2015.

\bibitem{muxfeldt2014kinesthetic}
A.~Muxfeldt, J.-H. Kluth, and D.~Kubus, ``Kinesthetic teaching in assembly
  operations--a user study,'' in {\em International Conference on Simulation,
  Modeling, and Programming for Autonomous Robots (SIMPAR)}, pp.~533--544,
  Springer, 2014.

\bibitem{akgun2012teleop}
B.~Akgun, K.~Subramanian, and A.~L. Thomaz, ``Novel interaction strategies for
  learning from teleoperation,'' in {\em AAAI Fall Symposium: Robots Learning
  Interactively from Human Teachers}, vol.~12, p.~7, 2012.

\bibitem{muxfeldt2017assembly}
A.~Muxfeldt, S.~Gopinathan, T.~Coenders, and J.~Steil, ``A user study on
  human-robot-interactive recovery for industrial assembly problems,'' in {\em
  2017 26th IEEE International Symposium on Robot and Human Interactive
  Communication (RO-MAN)}, pp.~824--830, August 2017.

\bibitem{pervez2017teleop}
A.~Pervez, A.~Ali, J.~Ryu, and D.~Lee, ``Novel learning from demonstration
  approach for repetitive teleoperation tasks,'' in {\em 2017 IEEE World
  Haptics Conference (WHC)}, pp.~60--65, June 2017.

\bibitem{card1990designspace}
S.~K. Card, J.~D. Mackinlay, and G.~G. Robertson, ``The design space of input
  devices,'' in {\em Proceedings of the SIGCHI Conference on Human Factors in
  Computing Systems}, CHI '90, pp.~117--124, March 1990.

\bibitem{card1991designspace}
S.~K. Card, J.~D. Mackinlay, and G.~G. Robertson, ``A morphological analysis of
  the design space of input devices,'' {\em ACM Transactions on Information
  Systems}, vol.~9, pp.~99--122, April 1991.

\bibitem{fischer1990teleop}
P.~Fischer, R.~Daniel, and K.~V. Siva, ``Specification and design of input
  devices for teleoperation,'' in {\em 1990 IEEE International Conference on
  Robotics and Automation}, vol.~1, pp.~540--545, May 1990.

\bibitem{massie94thephantom}
T.~H. Massie and J.~K. Salisbury, ``The phantom haptic interface: A device for
  probing virtual objects,'' in {\em Proceedings of the ASME Dynamic Systems
  and Control Division}, pp.~295--301, 1994.

\bibitem{kessler1995cyberglove}
G.~D. Kessler, L.~F. Hodges, and N.~Walker, ``Evaluation of the cyberglove as a
  whole-hand input device,'' {\em ACM Transactions on Computer-Human
  Interaction}, vol.~2, pp.~263--283, Dec. 1995.

\bibitem{riek12woz}
L.~D. Riek, ``{Wizard of Oz Studies in HRI}: A systematic review and new
  reporting guidelines,'' {\em Journal of Human-Robot Interaction}, vol.~1,
  pp.~119--136, July 2012.

\bibitem{Bajcsy2018a}
A.~Bajcsy, D.~P. Losey, M.~K. O'Malley, and A.~D. Dragan, ``{Learning from
  Physical Human Corrections, One Feature at a Time},'' in {\em 2018 ACM/IEEE
  International Conference on Human-Robot Interaction (HRI)}, pp.~141--149, ACM
  Press, March 2018.

\bibitem{kober13rl}
J.~Kober, J.~A. Bagnell, and J.~Peters, ``Reinforcement learning in robotics: A
  survey,'' {\em The International Journal of Robotics Research}, vol.~32,
  no.~11, pp.~1238--1274, 2013.

\bibitem{kabir2016primitives}
A.~Mohseni-Kabir, S.~Chernova, and C.~Rich, ``Identifying reusable primitives
  in narrated demonstrations,'' in {\em 2016 11th ACM/IEEE International
  Conference on Human-Robot Interaction (HRI)}, pp.~479--480, March 2016.

\bibitem{konidaris12cst}
G.~Konidaris, S.~Kuindersma, R.~Grupen, and A.~Barto, ``Robot learning from
  demonstration by constructing skill trees,'' {\em The International Journal
  of Robotics Research}, vol.~31, no.~3, pp.~360--375, 2012.

\bibitem{Huang2018}
Y.~Huang and Y.~Sun, ``A dataset of daily interactive manipulation,'' tech.
  rep., ArXiv, July 2018.

\bibitem{vogt17human}
D.~Vogt, S.~Stepputtis, S.~Grehl, B.~Jung, and H.~B. Amor, ``A system for
  learning continuous human-robot interactions from human-human
  demonstrations,'' in {\em 2017 IEEE International Conference on Robotics and
  Automation}, pp.~2882--2889, May 2017.

\bibitem{bremner2016gestures}
P.~Bremner and U.~Leonards, ``Iconic gestures for robot avatars, recognition
  and integration with speech,'' {\em Frontiers in Psychology}, vol.~7, p.~183,
  2016.

\bibitem{Grollman2012}
D.~H. Grollman and A.~G. Billard, ``Robot learning from failed
  demonstrations,'' {\em International Journal of Social Robotics}, vol.~4,
  pp.~331--342, November 2012.

\bibitem{calinon13extra}
S.~Calinon, T.~Alizadeh, and D.~G. Caldwell, ``On improving the extrapolation
  capability of task-parameterized movement models,'' in {\em 2013 IEEE/RSJ
  International Conference on Intelligent Robots and Systems}, pp.~610--616,
  November 2013.

\bibitem{lee2015force}
A.~X. Lee, H.~Lu, A.~Gupta, S.~Levine, and P.~Abbeel, ``Learning force-based
  manipulation of deformable objects from multiple demonstrations,'' in {\em
  2015 IEEE International Conference on Robotics and Automation}, pp.~177--184,
  May 2015.

\bibitem{rozo16gmm}
L.~Rozo, S.~Calinon, D.~G. Caldwell, P.~Jiménez, and C.~Torras, ``Learning
  physical collaborative robot behaviors from human demonstrations,'' {\em IEEE
  Transactions on Robotics}, vol.~32, pp.~513--527, June 2016.

\bibitem{Luo2012}
J.~Luo and K.~Hauser, ``Interactive generation of dynamically feasible robot
  trajectories from sketches using temporal mimicking,'' in {\em 2012 IEEE
  International Conference on Robotics and Automation}, pp.~3665--3670, May
  2012.

\bibitem{Rakita2018rss}
D.~Rakita, B.~Mutlu, and M.~Gleicher, ``{RelaxedIK}: Real-time synthesis of
  accurate and feasible robot arm motion,'' in {\em Robotics: Science and
  Systems}, 2018.

\bibitem{Luo2015}
J.~Luo and K.~Hauser, ``Robust trajectory optimization under frictional contact
  with iterative learning,'' in {\em Robotics: Science and Systems}, 2015.

\bibitem{Dragan2011}
A.~D. Dragan, N.~D. Ratliff, and S.~S. Srinivasa, ``Manipulation planning with
  goal sets using constrained trajectory optimization,'' in {\em 2011 IEEE
  International Conference on Robotics and Automation}, pp.~4582--4588, May
  2011.

\bibitem{Kalakrishnan2011}
M.~Kalakrishnan, S.~Chitta, E.~Theodorou, P.~Pastor, and S.~Schaal, ``{STOMP}:
  Stochastic trajectory optimization for motion planning,'' in {\em 2011 IEEE
  International Conference on Robotics and Automation}, pp.~4569--4574, May
  2011.

\bibitem{Zucker2013}
M.~Zucker, N.~Ratliff, A.~D. Dragan, M.~Pivtoraiko, M.~Klingensmith, C.~M.
  Dellin, J.~A. Bagnell, and S.~S. Srinivasa, ``{CHOMP: Covariant Hamiltonian
  optimization for motion planning},'' {\em The International Journal of
  Robotics Research}, vol.~32, pp.~1164--1193, September 2013.

\bibitem{guru2018icra}
G.~Subramani, M.~Gleicher, and M.~Zinn, ``Recognizing geometric constraints in
  human demonstrations using force and position signals,'' {\em IEEE Robotics
  and Automation Letters}, vol.~3, pp.~1252--1259, April 2018.

\bibitem{haage2017teaching}
M.~Haage, G.~Piperagkas, C.~Papadopoulos, I.~Mariolis, J.~Malec, Y.~Bekiroglu,
  M.~Hedelind, and D.~Tzovaras, ``Teaching assembly by demonstration using
  advanced human robot interaction and a knowledge integration framework,''
  {\em Procedia Manufacturing}, vol.~11, pp.~164--173, 2017.

\bibitem{lin2015remote}
H.~Lin, T.~Tang, Y.~Fan, Y.~Zhao, M.~Tomizuka, and W.~Chen, ``Robot learning
  from human demonstration with remote lead through teaching,'' in {\em 2016
  European Control Conference (ECC)}, pp.~388--394, June 2016.

\bibitem{orendt2016oneshot}
E.~M. Orendt, M.~Fichtner, and D.~Henrich, ``Robot programming by non-experts:
  Intuitiveness and robustness of one-shot robot programming,'' in {\em 2016
  25th IEEE International Symposium on Robot and Human Interactive
  Communication (RO-MAN)}, pp.~192--199, August 2016.

\bibitem{rakita2018camera}
D.~Rakita, B.~Mutlu, and M.~Gleicher, ``{An Autonomous Dynamic Camera Method
  for Effective Remote Teleoperation},'' in {\em 2018 ACM/IEEE International
  Conference on Human-Robot Interaction (HRI)}, pp.~325--333, ACM Press, March
  2018.

\bibitem{haptic-survey}
B.~Hannaford and A.~M. Okamura, ``{Haptics},'' in {\em Springer Handbook of
  Robotics}, pp.~719--739, Berlin, Heidelberg: Springer Berlin Heidelberg,
  2008.

\bibitem{davis1989perceived}
F.~D. Davis, ``Perceived usefulness, perceived ease of use, and user acceptance
  of information technology,'' {\em MIS Quarterly}, vol.~13, pp.~319--340,
  September 1989.

\bibitem{hart1988development}
S.~G. Hart and L.~E. Staveland, ``Development of {NASA-TLX (Task Load Index)}:
  Results of empirical and theoretical research,'' in {\em Human Mental
  Workload} (P.~A. Hancock and N.~Meshkati, eds.), vol.~52 of {\em Advances in
  Psychology}, pp.~139--183, North-Holland, 1988.

\bibitem{Kerracher2017}
N.~Kerracher and J.~Kennedy, ``Constructing and evaluating visualisation task
  classifications: Process and considerations,'' {\em Computer Graphics Forum},
  vol.~36, no.~3, pp.~47--59, 2017.

\bibitem{Guru2018corl}
G.~Subramani, M.~Zinn, and M.~Gleicher, ``Inferring geometric constraints in
  human demonstrations,'' in {\em Proceedings of The 2nd Conference on Robot
  Learning}, vol.~87 of {\em Proceedings of Machine Learning Research},
  pp.~223--236, PMLR, October 2018.

\bibitem{Guru2019icra}
G.~Subramani, M.~Zinn, and M.~Gleicher, ``{Robust replay of human
  demonstrations using identified constraints},''
\newblock {Submitted for publication}.

\end{thebibliography}
\end{document}